\documentclass{article}

\usepackage{microtype}
\usepackage{amsmath}
\usepackage{amsfonts}

\usepackage{graphicx}
\usepackage{subcaption}
\usepackage{booktabs} 
\usepackage{algorithm}
\usepackage{algpseudocode}

\usepackage{hyperref}

\usepackage[accepted]{mlsys2021_vrl}

\mlsystitlerunning{Sequantial Aggregation and Rematerialization for Training Large GNNs}

\begin{document}

\twocolumn[
\mlsystitle{Sequential Aggregation and Rematerialization: Distributed Full-batch Training of Graph Neural Networks on Large Graphs}




\begin{mlsysauthorlist}
\mlsysauthor{Hesham Mostafa}{Intel}
\end{mlsysauthorlist}

\mlsysaffiliation{Intel}{Intel AI Lab, San Diego, California, USA}

\mlsyscorrespondingauthor{Hesham Mostafa}{hesham.mostafa@intel.com}

\mlsyskeywords{Graph Neural Networks, Distributed Training, Rematerialization, Large-scale Training}

\vskip 0.3in

\begin{abstract}
  We present the Sequential Aggregation and Rematerialization (SAR) scheme for distributed full-batch training of Graph Neural Networks (GNNs) on large graphs. Large-scale training of GNNs has recently been dominated by sampling-based methods and methods based on non-learnable message passing.
  SAR on the other hand is a distributed technique that can train any GNN type directly on an entire large graph. The key innovation in SAR is the distributed sequential rematerialization scheme which sequentially re-constructs then frees pieces of the prohibitively large GNN computational graph during the backward pass. This results in excellent memory scaling behavior where the memory consumption per worker goes down linearly with the number of workers, even for densely connected graphs. Using SAR, we report the largest applications of full-batch GNN training to-date, and demonstrate large memory savings as the number of workers increases. We also present a general technique based on kernel fusion and attention-matrix rematerialization to optimize both the runtime and memory efficiency of attention-based models. We show that, coupled with SAR, our optimized attention kernels lead to significant speedups and memory savings in attention-based GNNs. We made the SAR GNN training library publicy available: \url{https://github.com/IntelLabs/SAR}.
  
\end{abstract}
]



\printAffiliationsAndNotice{} 

\section{Introduction}
\label{sec:intro}
Graph neural networks (GNNs), also called message passing networks~\cite{Gilmer_etal17,Kipf_Welling16,Scarselli_etal08}, are widely applied in various types of graph-related problems~\cite{Shi_etal20,Corso_etal20,Li_etal19,Zhang_Chen18}. At every layer in a GNN, each node produces an output feature vector by using a learnable transformation to aggregate the input feature vectors of its neighbors in the graph. After $K$ layers, a node's receptive field would span the input features of all nodes that are less than $K$ hops away in the graph. While this could be advantageous in order to allow a GNN to consider a wider neighborhood when learning the features of each node~\cite{Li_etal20}, it can lead to a large computational graph at the output where every node's output features depend on a significant portion of the entire input graph, as well as intermediate node features, a phenomenon that is known as neighbor explosion~\cite{Hamilton_etal17}.

When training deep GNNs on large graphs, it can quickly becomes infeasible to store the large GNN computational graph in memory. Exact GNN training, or full-batch GNN training, is thus difficult to scale to large graphs. Other, more scalable, training alternatives such as sampling-based methods~\cite{Hamilton_etal17,Zeng_etal19,Chiang_etal19} have recently become more common. Sampling-based methods keep the memory requirements in check by sampling a small part of the graph each training iteration. Due to the sampling involved, these approaches yield noisy and biased gradient estimates~\cite{Chen_etal18,Chen_etal18a}. Many sampling-based methods run into memory issues as the depth of the GNN increases, and bigger neighborhoods need to be sampled for each node~\cite{Hamilton_etal17}. The sampling operation itself introduces extra computational overhead. With the wide variety of possible sampling operators, it is unclear for a given problem what sampling operator would work best. While it is possible to bound the errors introduced by sampling in linear GNN layers~\cite{Chen_etal18}, no such bounds exist when the GNN uses non-linearities. Many scalable GNN training methods avoid having to construct the large computational graph by using non-learnable message propagation followed by learnable node-wise networks
~\cite{Rossi_etal20,Sun_Wu21,Klicpera_etal18,Wu_etal19}. This simplifies the GNN considerably as the expensive propagation of messages between neighbors in the graph is only done once in a non-learnable manner during pre-processing. The use of non-learnable messages, however, limits the expressiveness of these models compared to traditional GNNs.

\begin{figure*}[h]
  \centering
  \begin{subfigure}{0.48\textwidth}
    \includegraphics[width = \textwidth]{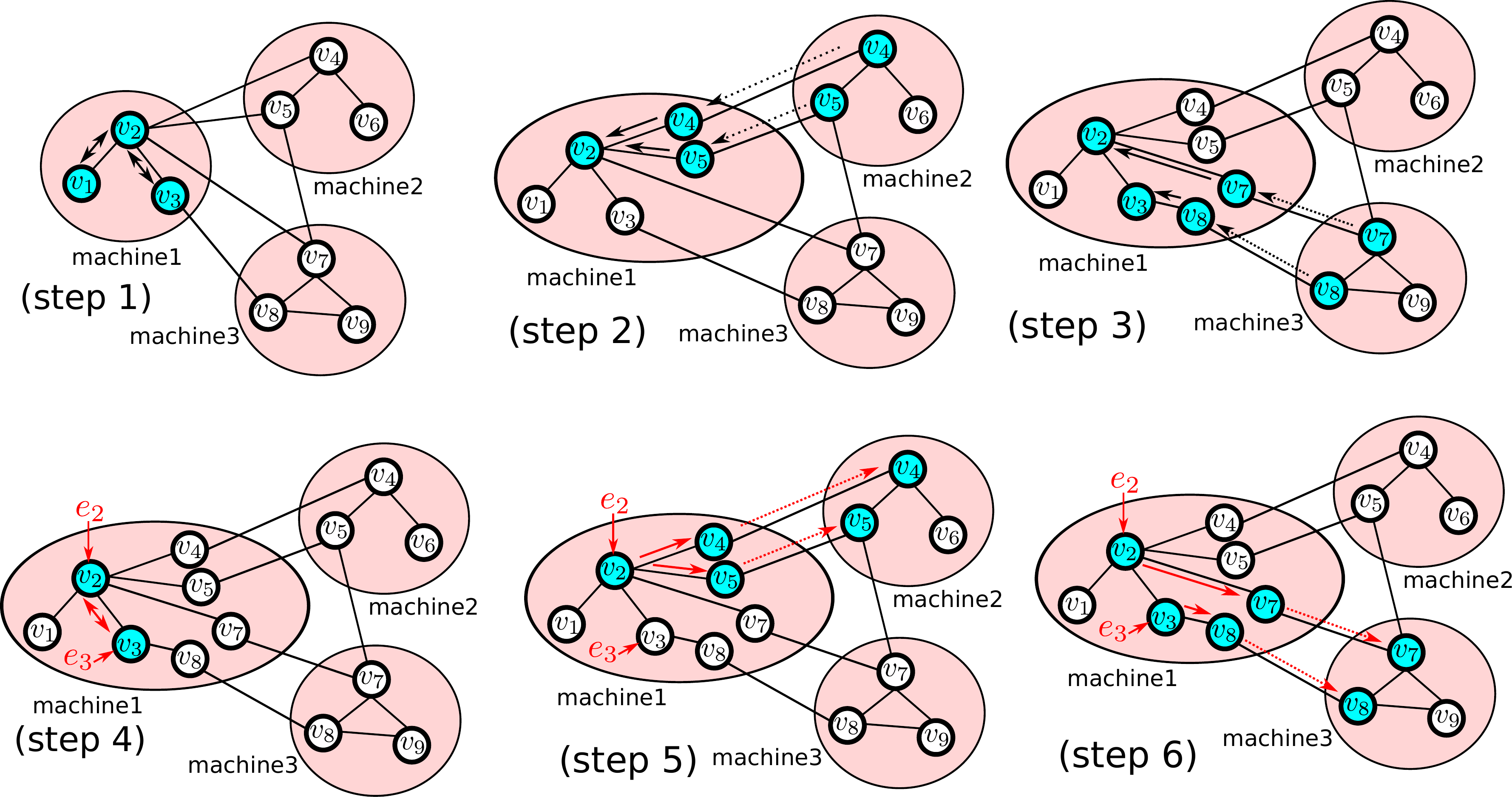} 
    \subcaption{}
    \label{fig:dom_parallel_a}
  \end{subfigure}
  \quad
  \begin{subfigure}{0.48\textwidth}
    \includegraphics[width = \textwidth]{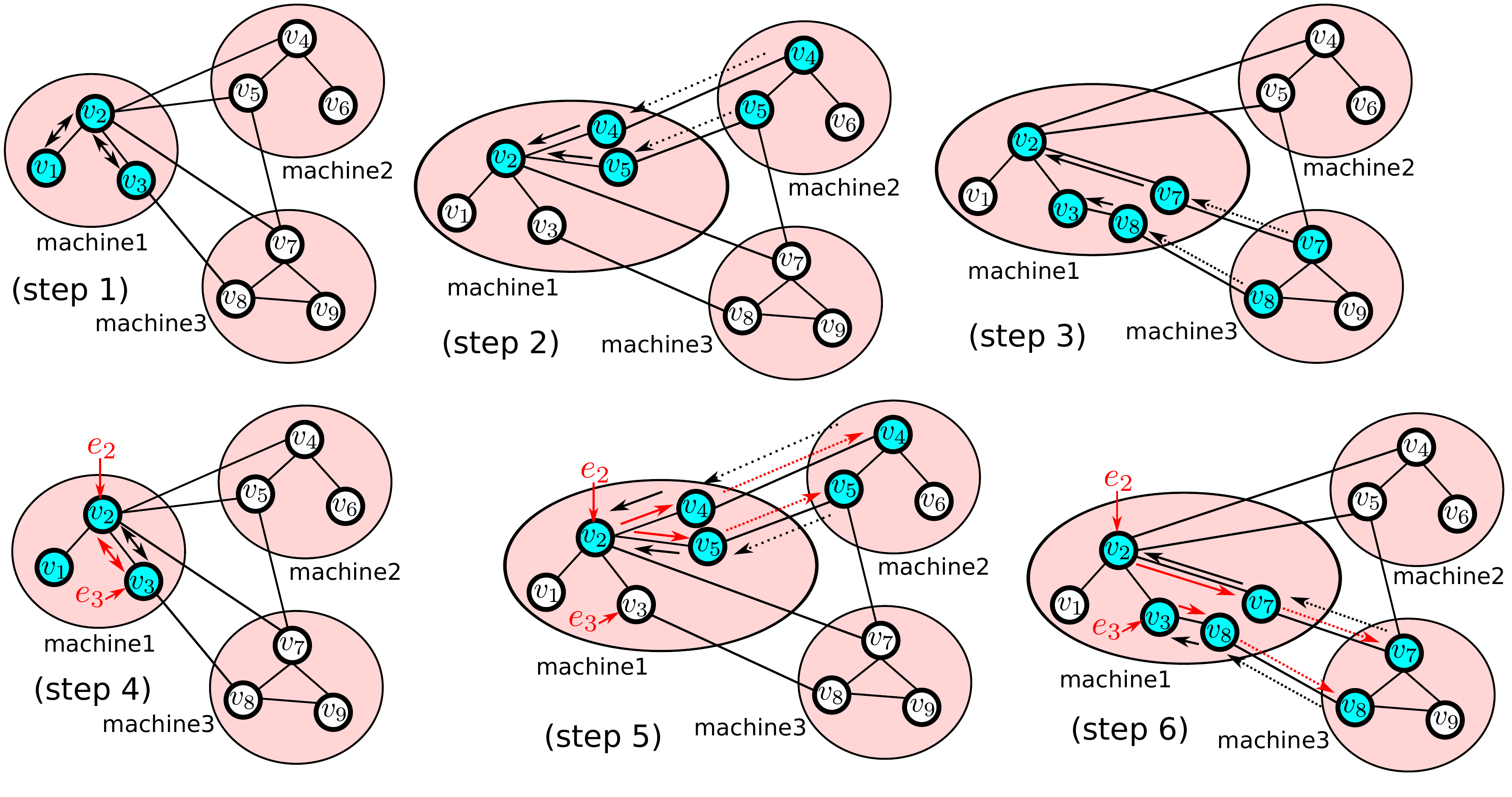} 
    \subcaption{}    
    \label{fig:dom_parallel_b}
  \end{subfigure}

  \caption{Domain parallel training of a GNN layer on a graph with 9 nodes. The graph is partitioned across 3 machines. Dashed arrows indicate inter-machine communication, while solid arrows indicate within machine message passing. Only the data flow and storage patterns for machine 1 are shown. (\subref{fig:dom_parallel_a}) Forward pass (top) and backward pass (bottom) for vanilla domain-parallel training. Note how at the end of the forward pass (step 3), machine 1 needs to store the input features of 7 nodes as part of the output's computational graph. (\subref{fig:dom_parallel_b}) Forward pass (top) and backward pass (bottom) for domain-parallel training using SAR. The computational graph is not constructed during the forward pass. Instead, it is sequentially rematerialized then deleted during the backward pass.}
  \label{fig:dom_parallel}
\end{figure*}

Distributed training across several machines might seem to be one way to handle the large memory requirements of full-batch GNN training. However, distributed training methods such as model parallel training would still run into issues if a single device cannot accommodate the input graph or the activations of a single GNN layer. A more promising approach is domain parallel training~\cite{Gholami_etal18}, also known as spatially-parallel training in the convolutional neural networks case~\cite{Dryden_etal19,Jin_etal18}. In domain-parallel training, the input is split into many parts and each machine handles the computation for a single part. Figure~\ref{fig:dom_parallel_a}, however, illustrates the issue with domain-parallel training when applied to GNNs. Even though, initially, each machine stores only a small part of the input graph, each machine would eventually need to store a substantial portion of the entire graph as part of its output's computational graph.

Our main contribution is the sequential aggregation and rematerialization (SAR) scheme illustrated in Fig.~\ref{fig:dom_parallel_b}. SAR avoids constructing the computational graph during the forward pass. Instead, SAR constructs the computational graph and frees it piece by piece during the backward pass. Assuming equal partitioning of the graph across $N$ workers, we can show that SAR would need to materialize at most 2 graph partitions at each machine at any point in time. The memory requirement per machine thus scales as $2/N$, even for densely connected graphs. This allows SAR to scale to arbitrarily large graphs by simply adding more workers.

We show that the communication overhead incurred by SAR to re-materialize the computational graph during the backward pass can be avoided for many popular GNN variants, making the memory savings of SAR practically free. For some variants such as Graph Attention Networks (GATs)~\cite{Velivckovic_etal17}, however, the communication overhead of SAR can not be avoided. Yet, we identify a couple of optimizations for attention-based models which synergize particularly well with SAR. These optimizations avoid materializing the costly attention coefficients tensors and instead compute them on the fly using fused kernels during the forward and backward passes. We show this speeds up the computation in GAT-like networks by reducing redundant memory accesses, and in conjunction with SAR, further reduces the memory footprint for attention-based models. We show that after incorporating these optimizations, training GAT using SAR is as fast as vanilla domain parallel training while consuming a fraction of the memory.

We build SAR directly on top of DGL~\cite{Wang_etal19}, one of the most popular GNN training libraries, which in turn is built on top of PyTorch~\cite{Paszke_etal19}. This allows us to directly use standard DGL layers and GNN networks. Using SAR requires minor modification to existing single-node DGL code: it requires modifying the data loading part to load a graph partition at each worker instead of loading the entire graph; and modifying the training loop to synchronize the parameter gradients at the end of each training iteration. Unlike prior full-batch GNN training frameworks that reimplement basic graph operations~\cite{Ma_etal19,Jia_etal20}, we directly leverage the graph operations of DGL which allows us to capitalize on efficient, continuously updated, kernels for many basic graph operations such as sparse-dense matrix multiplications (SpMM). We integrate the sequential rematerialization steps of SAR into PyTorch's Autograd mechanics in a way that is transparent to the user. The user can thus use PyTorch's standard model definition steps to describe arbitrary GNN topologies. During training, SAR runs under the hood to manage inter-machine communication and the dynamic construction and deletion of pieces of the computational graph. We test SAR on a cluster of Xeon CPUs and use it to run the largest full-batch GNN training experiments to date. We report competitive accuracies on all benchmarks while using a fraction of the memory consumed by standard domain parallel training.

\section{Related Work}
\label{sec:related}
Memory is often the central bottleneck when running full-batch graph training. NeuGraph~\cite{Ma_etal19} describe an efficient hybrid CPU-GPU training method that continuously shuttles data between main memory and GPU memory to enable full-batch GNN training on the GPU. NeuGraph, however is limited to multi-GPU training on a single host where the graph and GNN activations need to fit in system memory. ROC~\cite{Jia_etal20} removes the single-node limitation to enable multi-node training, yet both NeuGraph and ROC train on relatively small graphs that easily fit within the main memory of a single host. They do not address scalability concerns for truly large-scale (multi-terabyte) GNN models that exceed the realistic memory capacity of a single host. CAGNET~\cite{Tripathy_etal20} describes a distributed training method that minimizes inter-node communication. However, it still tests on relatively small graphs and does not address memory scalability concerns. All these methods do not tackle  complex GNN models like GAT. They are also specific to linear GNN topologies and do not address more complex topologies that make use of skip connections~\cite{Li_etal20,Xu_etal18}. DistGNN~\cite{Md_etal21} is a recent distributed GNN training library for full-batch training. However, `full-batch'  in DistGNN refers only to the forward pass. The backpropagation of errors is local to each machine and no inter-machine communication happens during the backward pass leading to a degradation of accuracy as the number of machines increases.
SAR runs a full-batch forward {\it and} backward pass. The results of training are exactly the same regardless of the number of machines.

Activation re-materialization is a popular strategy for reducing the training memory footprint of deep learning models. Activation materialization avoids storing all intermediate activations during the forward pass. Instead, during the backward pass, it recomputes many intermediate activations or loads them directly from disk~\cite{Chen_etal16,Jain_etal19}. Activation re-materialization differs from SAR in that SAR maintains all activations in the memory of the hosts. SAR does not recompute or load activations from disk, but rather shuttles activations between hosts to wherever they are needed to compute outputs during the forward pass, or to compute gradients during the backward pass. SAR guarantees that this shuttling of activations will never overload the memory of a single host with more than two partitions of the data at any point in time.

Reversible neural networks use special layers that allows a layer's input to be reconstructed (rematerialized) from the layer's output during the backward pass~\cite{Gomez_etal17,Li_etal21}. While the training memory footprint of reversible networks can be made independent of the number of layers, reversible networks incur an additional computational cost to recompute activations. They also require a single layer's computational graph to fit in memory, which might not be possible for GNNs applied to large graphs.

\section{Methods}
\subsection{Preliminaries}
Let $G(\mathcal{V},\mathcal{E})$ be a graph with node set $\mathcal{V}$ and edge set $\mathcal{E}$. Let $x_i\equiv h_i^0$ be the input feature vector of node $i$. Layer $l$ in a multi-layer GNN produces the output $h_i^l$ at node $i$ through the following steps.
\begin{flalign}
  z^{l} &= {\bf W}^l h^{l-1}, \quad m^{l}_{j\rightarrow i} = f_{message}(z_i^{l},z_j^{l};\theta^l), \nonumber  \\
  h_i^{l} &= f_{node}\left(h_i^{l-1},Agg\left(\{m^{l}_{j\rightarrow i} : j\in \mathcal{N}(i)\}\right);\phi^l\right). \label{eq:agg}
\end{flalign}
${\bf W}^l$ is a learnable matrix. $f_{message}$ and $f_{node}$ are general learnable transformations with parameters $\theta^l$ and $\phi^l$, respectively. $\mathcal{N}(i)$ is the neighborhood of node $i$ in the graph, and $m_{j\rightarrow i}$ are messages from node $j$ to an adjacent node $i$. $Agg$ is a permutation-invariant aggregation operator. We are interested in two representative variants of GNNs: GraphSage~\cite{Hamilton_etal17} and GAT~\cite{Velivckovic_etal17}. For GraphSage, the propagation and aggregation operators take the form:
\begin{flalign}
  m^{l}_{j\rightarrow i} &= z^{l}_{j}, \nonumber  \\
  h_i^{l} &= \sigma\left({\bf W}_{res}h_i^{l-1} + \frac{1}{|\mathcal{N}(i)|}\sum\limits_{j\in \mathcal{N}(i)}m^{l}_{j\rightarrow i}\right),  \label{eq:sage}
\end{flalign}
where ${\bf W}_{res}$ are learnable weights, and $\sigma$ is a non-linearity. Similarly, GAT is defined by:
\begin{flalign}
  e_{j\rightarrow i}^l &=  LeakyReLU\left(a^{l^T} (z^l_i || z_j^l)\right) \nonumber \\
  m^{l}_{j\rightarrow i} &= (e_{j\rightarrow i}^l,z_j^l) \nonumber \\
  h_i^{l} &= \sigma\left( \sum\limits_{j\in \mathcal{N}(i)}\frac{\exp(e_{j\rightarrow i}^l)}{\sum\limits_{k\in \mathcal{N}(i)}\exp(e_{k\rightarrow i}^l)} z_j^l\right),  \label{eq:gat}
\end{flalign}
 where $a^{l}$ is learnable. The message is a 2-tuple containing un-normalized attention weights and the neighbor features.

\subsection{Sequential Aggregation and Rematerialization}
\label{sec:sar}
The aggregation operator $Agg$ used in GNNs can typically be sequentially or recursively applied, i.e, $Agg(\{A,B,C\}) = Agg(\{A,Agg(\{B,C\})\})$. Even attention-based aggregators (such as in Eq.~\ref{eq:gat}) can be incrementally applied by sequentially aggregating neighboring features using un-normalized attention coefficients, and sequentially aggregating the normalization factors separately. At the end, we can divide by the normalization factors to obtain the correctly-normalized output. This property of the aggregation operator allows a machine in domain parallel training to sequentially fetch neighboring nodes from remote machines and aggregate their messages incrementally, and as we will show, also enables SAR to incrementally backpropagate the gradients along pieces of the computational graph. 

SAR begins by partitioning the graph into $N$ partitions and distributes these partitions to $N$ machines. ${\bf V}_p$ are the graph vertices in partition $p$. Each machine is responsible for computing and storing the features of the vertices in its partition for all GNN layers. For each worker $p$, SAR constructs $N$ subgraphs ${\mathcal G}_{p,1},\ldots,{\mathcal G}_{p,N}$ where ${\mathcal G}_{p,q} = ({\mathcal V}_{p,q},{\mathcal E}_{p,q})$. ${\mathcal E}_{p,q}$ are all the edges from the vertices in partition $q$ to the vertices in partition $p$, and ${\mathcal V}_{p,q}$ are the vertices incident to these edges.

The forward pass of SAR is described in algorithm 1. It uses standard PyTorch Autograd mechanics to record the computational trace until it reaches the point where it needs to execute the message passing and aggregation parts of the GNN layer. SAR then turns off PyTorch's Autograd, starts fetching remote data, and runs the messages construction and aggregation incrementally. SAR ensures that data from at most one remote partition is resident in memory at any time. After aggregation is complete, SAR re-enables PyTorch Autograd and the GNN's forward pass continues in a standard manner. There is thus a gap in the Autograd trace between the ${\bf z}^l$ and $\theta^l$ variables and the output of the $Agg$ operator: $Acc^l$ (see Eq.~\ref{eq:agg}).

During the backward pass, PyTorch Autograd can not backpropagate errors through this gap. SAR thus takes over as soon as it receives the error w.r.t (with respect to) the aggregator output ($e_{Acc^l}$) and handles the backpropagation of errors to ${\bf z}^l$ and $\theta^l$ as shown in Algorithm 2. SAR iterates over all partitions and incrementally backpropagates the error to each of them. Worker $p$ waits until it receives the errors for its local node features, ${\bf Z}_p^l$, from all other workers (Line 15), then provides this error to Autograd (Line 17) to continue error backpropagation. SAR distinguishes between two cases:
\begin{itemize}
\item {\bf case 1}: The gradient of the aggregator output w.r.t ${\bf z}^l$ and $\theta$ does not depend on the values of ${\bf z}^l$. SAR does not need to fetch the remote ${\bf z}^l$ variables and can instead directly send the error back to their machine. That is the case for GraphSage (Eq.~\ref{eq:sage}) where aggregation simply sums up the features of neighboring nodes and there is no $\theta$ parameter.
\item {\bf case 2}: When case 1 does not hold (which is the case for GAT). SAR must fetch the remote ${\bf z}^l$ variables to correctly evaluate the gradients.
\end{itemize}
These two cases reflect the two distinct behaviors of mathematical operators in PyTorch (as well as other machine learning libraries): in the first case, the operator {\it does not} need  the input tensors in order to backpropagate the gradients from the output to the inputs. The simplest example is the sum operator; in the second case the operator {\it needs} the input tensors in order to be able to calculate the gradients of the inputs from the output gradient in the backward pass. The simplest example is the product operator.

Since many GNN variants such as GraphSage use sum or mean aggregation, SAR will introduce no overhead compared to vanilla domain parallel training for these GNN variants. For GNN variants like GAT, there is an extra communication and computation overhead to re-fetch the remote features during the backward pass and re-calculate any intermediate variables (such as the attention coefficients $e^l_{j\rightarrow i}$ in Eq.~\ref{eq:gat}). The communication overhead in case 2 over vanilla domain parallel training is 50\% as instead of only communicating node features during the forward pass and  gradients of the node features during the backward pass, SAR also has to communicate node features during the backward pass.  Figure~\ref{fig:dom_parallel_b} illustrates the backward pass for case 2. In the next section, we describe how this overhead can be minimized.

\begin{algorithm}
  \caption{SAR forward pass for layer $l$ on worker $p$}
\begin{algorithmic}
\label{alg:forward}

  \State {\bf Inputs}: $\{h_k^{l-1} : k \in {\bf V}_p\},{\mathcal G}_{p,1},\ldots,{\mathcal G}_{p,N}$
  \State $z_k^{l} = {\bf W}^lh_k^{l-1}$ for all $k \in {\bf V}_p$
  \State Disable PyTorch Autograd
  \State $Acc^l = 0 \in \mathbb{R}^{|{\bf V}_p| \times F}$ \Comment{Output of the $Agg$ operator}
  \For{$q \in 1..N$} \Comment{Iterate over all  partitions}
  \State Fetch ${\bf Z}_{q\rightarrow p}^l=\{z_k^{l} : k \in {\bf V}_q \wedge k \in {\mathcal V}_{p,q}\}$
  \For{$(i,j) \in \mathcal{E}_{p,q}$}
  \State $m^{l}_{j\rightarrow i} = f_{message}(z_i^{l},z_j^{l};\theta^l)$ 
  \EndFor
  \For{$i \in {\bf V}_p \wedge i \in \mathcal{V}_{p,q}$}
  \State $Acc^l_i = Acc^l_i +  Agg\left(\{m^{l}_{j\rightarrow i} : (i,j)\in \mathcal{E}_{p,q}\}\right) $
  \EndFor
  \State delete ${\bf Z}_{q\rightarrow p}^l$ and  $\{m^{l}_{j\rightarrow i} : (i,j) \in \mathcal{E}_{p,q}\}$
  \EndFor
  \State Enable PyTorch Autograd
  \State {\bf Return} $h_i^{l} = f_{node}(h_i^{l-1},Acc^l;\phi^l)$

\end{algorithmic}
\end{algorithm}

\begin{algorithm}
  \caption{SAR backpropagation of errors within the message passing and aggregation part of layer  $l$ on worker $p$}
\begin{algorithmic}
\label{alg:backward}

  \State {\bf Inputs} : ${\bf Z}_p^l = \{z_k^{l} : k \in {\bf V}_p\},Acc^l,e_{Acc^l}$
  \State {\bf Inputs} : ${\mathcal G}_{p,1},\ldots,{\mathcal G}_{p,N}$
  \State $\theta^l.grad = 0$ \Comment{Gradient of theta parameter}
  \For{$q \in 1..N$} \Comment{Iterate over all  partitions}
  \If{$\frac{dAcc^l}{dz^{l}}$ or $\frac{dAcc^l}{d\theta^l}$ depend on $z^{l}$}
  \State Fetch ${\bf Z}_{q\rightarrow p}^l = \{z_k^{l} : k \in {\bf V}_q \wedge k \in {\mathcal V}_{p,q}\}$ 
  \EndIf
  \State $E_{p\rightarrow q} =  \{e^T_{Acc^l} \frac{dAcc^l}{dz_k^l} : k \in {\bf V}_q \wedge k \in {\mathcal V}_{p,q}\}$  
  \For{$(i,j) \in \mathcal{E}_{p,q}$}
  \State $\theta^l.grad = \theta^l.grad + e^T_{Acc^l}\frac{dAcc^l}{dm^l_{j\rightarrow i}}\frac{dm^l_{j\rightarrow i}}{d\theta^l} $ 
  \EndFor
  \State send error $E_{p\rightarrow q}$ to worker $q$
  \EndFor
  \State \Comment{Wait and accumulate errors from remote machines:}
  \State $E_p = \sum\limits_{q = 1}^N E_{q\rightarrow p}$  \Comment{Total Loss gradient w.r.t ${\bf Z}^l_p$}
  \State Sum $\theta^l.grad$ across all machines
  \State ${\bf Z}^l_p.backward(E_p)$ 
  
\end{algorithmic}
\end{algorithm}

\subsection{Fused Attention Kernels}
Models that make use of an attention mechanism such as transformers~\cite{Vaswani_etal17} or GAT~\cite{Velivckovic_etal17} are typically implemented in two steps: 1) Calculate the attention coefficients, 2) Sum the input features weighted by the attention coefficients. This two-step process writes the attention coefficients to memory after step 1 then reads them back in step 2. We implemented custom fused forward and backward kernels for GAT (Eq.~\ref{eq:gat}) that calculate the attention coefficients on the fly while summing the node's neighbor features (forward pass) or pushing the gradients to the neighbors (backward pass). The attention coefficients are thus never written to, or read from, memory. This reduces memory accesses and reduces the peak memory consumption of the fused kernel compared to the standard DGL implementation. One downside when running standard training is that we need to re-calculate the attention coefficients during the backward pass. This downside, however, does not apply when we do distributed training using SAR, as we would need to re-materialize the computational graph during the backward pass anyway, which involves re-calculating the attention coefficients. Using fused attention kernels thus synergizes well with SAR and as we will show in the next section significantly speeds up distributed training of GAT networks.

As we show in the results section, we find that the fused forward and backward attention kernels often outperform DGL's vanilla GAT implementation in vanilla single-host training. The benefits of the reduced memory access profile of the fused kernels thus outweighs the cost of re-calculating the attention coefficients in the backward pass.

\subsection{Practical Considerations}
\paragraph{Prefetching:} SAR is able to guarantee that no more than two graph partitions will be resident in memory at any worker. However, to overlap computation and communication, we find it beneficial to use separate pre-fetching threads to pre-fetch the next remote partition at each worker before the worker is done with the current partition. There can thus be 3 partitions resident in memory: the local partition, the remotely-fetched partition whose messages are being actively aggregated, and the partition being pre-fetched. We still maintain linear per-worker memory scaling with the number of workers $N$, but at a rate of $3/N$ instead of $2/N$.

\paragraph{Stable softmax:} The softmax operation is often used to normalize attention coefficients: $softmax({\bf e})_i=\exp(e_i)/\sum\limits_k \exp(e_k)$. Standard implementations first subtract the maximum from all softmax arguments to avoid large exponentials. In SAR's incremental aggregation, we do not have access to the maximum of the softmax arguments. We thus use a running estimate of the maximum based on the components of ${\bf e}$ we have observed till now. Whenever we update the maximum, we correct the accumulated numerator and denominator by multiplying with $\exp(old\_max - new\_max)$. We observe that training can quickly become unstable if we fail to numerically stabilize the softmax using this method.

\paragraph{Batch normalization:} Batch normalization normalizes each row in the node feature matrix ${\bf H}\in \mathbb{R}^{|V| \times F}$ to yield $\hat{{\bf H}}$, where $\hat{{\bf H}}[i] = ({\bf H}[i] - \mu_B)/\sqrt{\sigma_B^2 + \epsilon}$.  $\mu_B$ and $\sigma_B$ are the mean and variance calculated across all rows of ${\bf H}$. In SAR, ${\bf H}$ is distributed across the workers. SAR implements a custom BatchNorm operation that collectively calculates the global mean and variance from the local mean and variance in each worker, and a custom backward pass that ensures the gradients w.r.t $\mu_B$ and $\sigma_B$ are properly scaled based on the number of nodes(feature matrix rows) at each workers, before backpropagating further to the feature matrix. By communicating only summary statistics and their gradients, SAR's BatchNorm is scalable and communication-efficient.

\section{Experimental Results}
We run all experiments on a cluster of machines with Xeon CPUs where each CPU has 36 physical cores, and each machine has 256GB of main memory. The machines are connected through Mellanox Infiniband 200Gb/s HDR interconnects. We use Python 3.6.8, PyTorch 1.8.1, and  DGL 0.7. We compile our custom GAT kernels using Intel's icc compiler.  All inter-machine communication is handled by the torch.distributed library, where we use torch\_ccl~\cite{torch_ccl} as the communication backend. torch\_ccl is the PyTorch wrapper for Intel's OneCCL collective communication library~\cite{oneccl}.
\subsection{Single-host Performance of Fused Attention Kernels}
\begin{figure}[h]
  \centering
  \begin{subfigure}{0.225\textwidth}
    \includegraphics[width = \textwidth]{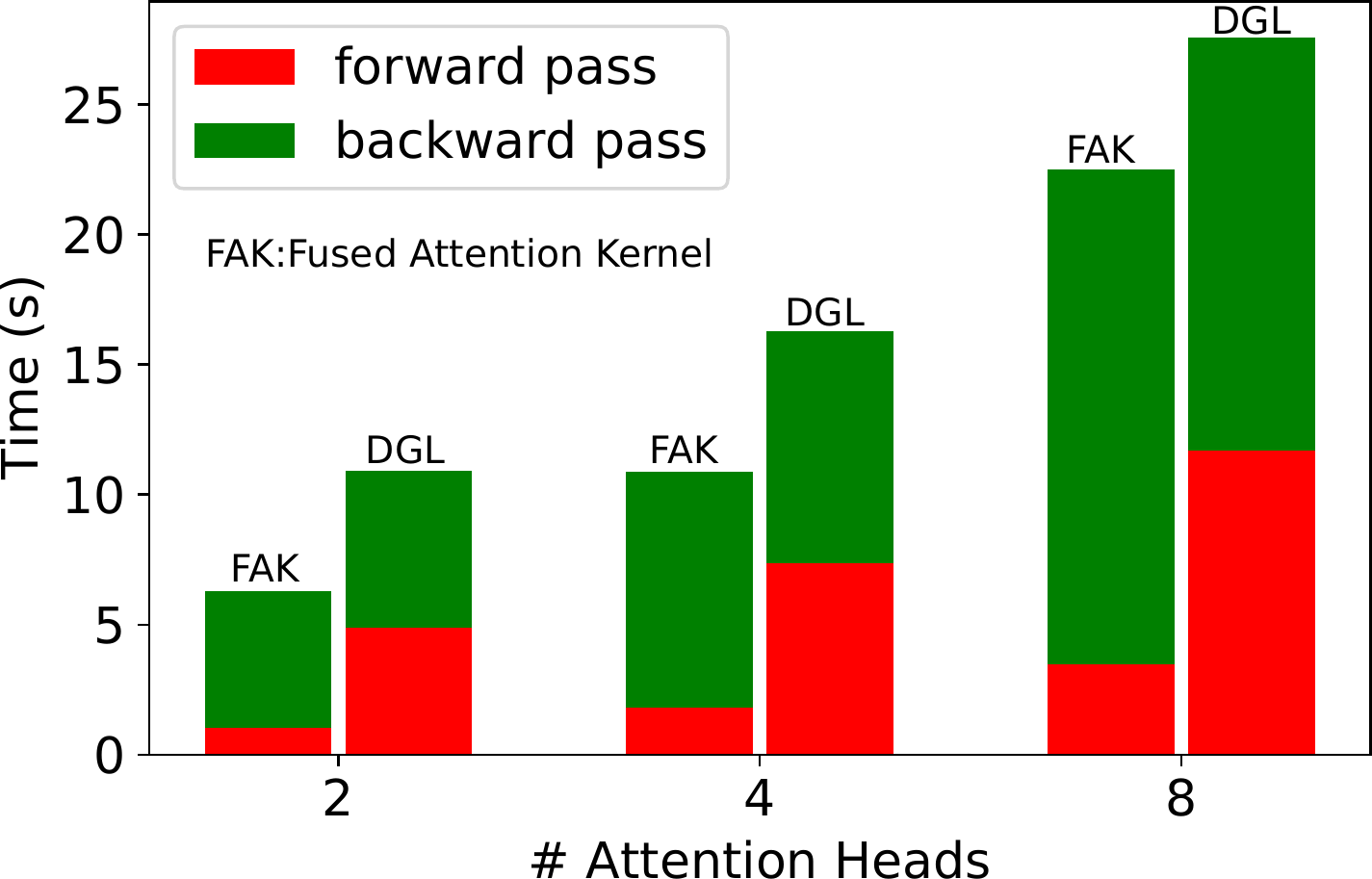} 
    \subcaption{}
    \label{fig:FAK_runtime}
  \end{subfigure}
  \quad
  \begin{subfigure}{0.225\textwidth}
    \includegraphics[width = \textwidth]{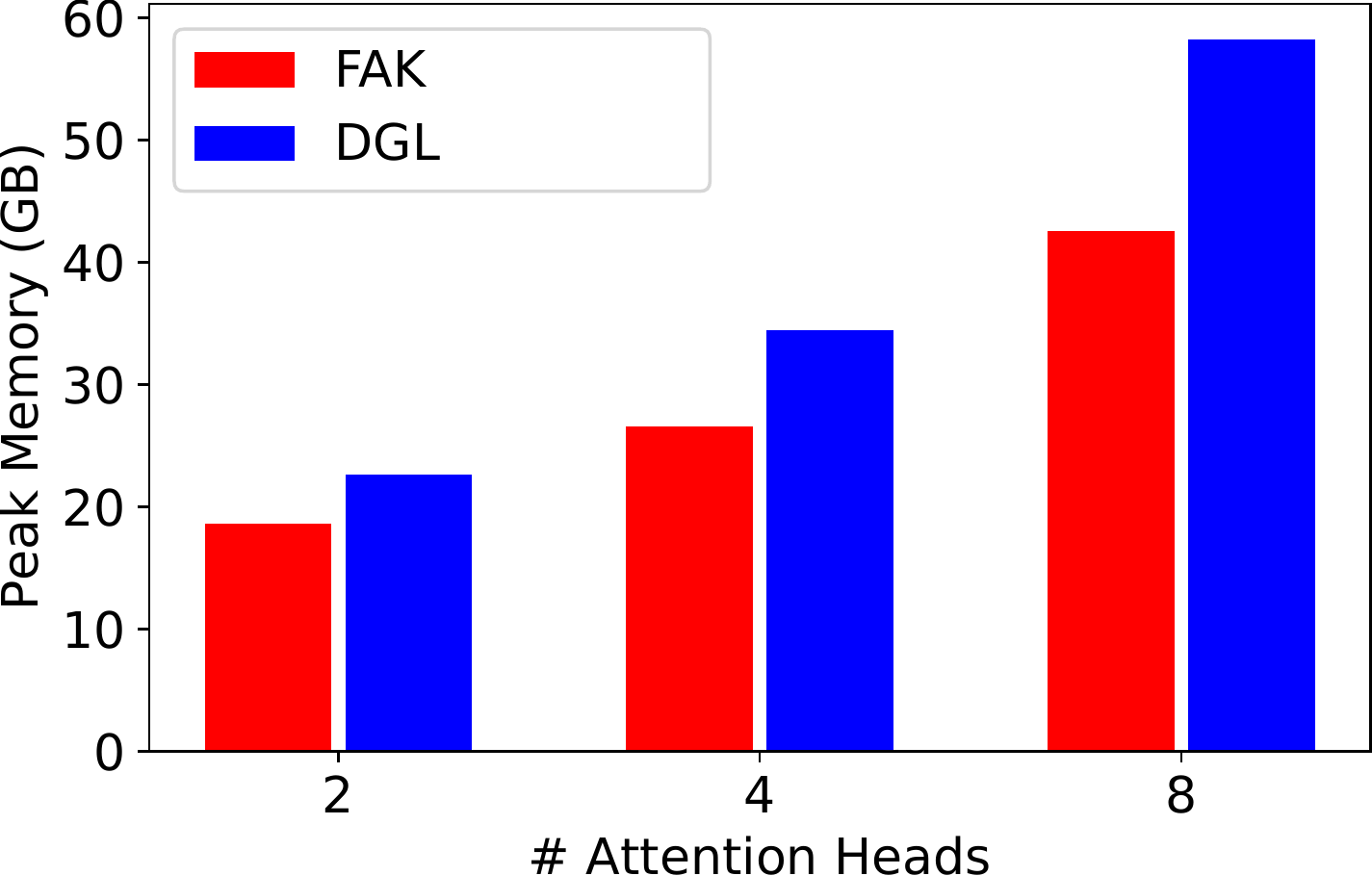} 
    \subcaption{}    
    \label{fig:FAK_memory}
  \end{subfigure}

  \caption{(\subref{fig:FAK_runtime}) Run times for the forward and backward passes for the fused attention kernel (FAK) and DGL's GAT implementation. (\subref{fig:FAK_memory}) Peak memory consumption for the FAK and DGL's GAT implementation.}
  \label{fig:FAK}
\end{figure}

We compare the performance of our fused attention kernel for GAT against the standard DGL implementation of GAT layers. We use the ogbn-products graph (see Table 1) and measure the run-time and memory consumption for the forward and backward passes of a single GAT layer. We vary the number of attention heads between 2,4, and 8  while keeping the feature dimension per head constant which varies the input and output feature dimensions between 200, 400, and 800. The number of attention heads controls how many attention coefficients need to be calculated per edge. Figure~\ref{fig:FAK_runtime} shows that the forward pass for the fused kernel is much faster than DGL (up to 4.5x faster for 2 attention heads). We see that the fused backward kernel is slightly faster than DGL's backward pass when using 2 heads, is on par with DGL's backward pass when using 4 heads, and slower when using 8 heads. As we increase the number of attention heads, the fused backward kernel needs to re-calculate more attention coefficients per edge during the backward pass, which explains its worsening relative performance as the number of heads increases. The combined forward+backward pass using the fused kernels is, however, always faster then DGL, though the advantage narrows as we increase the number of heads. 

Figure~\ref{fig:FAK_memory} shows the peak memory consumption (at the end of the forward pass). As expected, the fused kernel has lower peak memory as it does not store the attention coefficients during the forward pass. The memory advantage of the fused kernel increases as we increase the number of attention heads (which increase the number of attention coefficients).

\subsection{SAR Performance on Large-scale Graphs}
\begin{table}[h]
\caption{Graph Datasets}
\label{tab:dataset_properties}
\begin{center}
\begin{tabular}{l|cc}
\hline
  & \begin{tabular}{@{}c@{}}ogbn- \\ products\end{tabular} & \begin{tabular}{@{}c@{}}ogbn- \\ papers100M\end{tabular} \\
    \hline
    \hline
    \# nodes & 2.5M & 111M \\
    \# edges & 124M & 3.2B \\    
    \# input features & 100 & 128 \\
    \# classes & 47 & 172\\
    GraphSage Accuracy & 80.1\% & 65.8\% \\
    GraphSage+C\&S Accuracy & 80.9\% & 66.3\% \\    
    GAT Accuracy & 74.9\% & 63.7\% \\
    GAT+C\&S Accuracy & 77.7\% & 64.5\% \\    
\end{tabular}
\end{center}
\end{table}

We test SAR on two of the largest graphs in the Open Graph Benchmarks(OGB) datasets~\cite{Hu_etal20}: ogbn-products and ogbn-papers100M. These are homogeneous graphs.  Appendix~\ref{sec:rgcn} contains results on a heterogeneous graph (ogbn-mag). Table~\ref{tab:dataset_properties} summarizes the properties of ogbn-products and ogbn-papers100M. We train two GNNs on each graph: 1) a 3-layer GraphSage network with hidden feature size of 256, 2) a 3-layer GAT network with hidden feature size of 128 and 4 attention heads. We use batch normalization and dropout between all layers. We use the label augmentation and masked label prediction scheme from ref.~\cite{Shi_etal20} where, each epoch,  we augment the input features of a random subset of the training nodes with the ground truth label, and predict the labels of the remaining training nodes. At inference time, we augment all training nodes with the ground truth labels. We always train for 100 epochs with a decaying learning rate. At the end of training, we run the Correct and Smooth (C\&S) procedure from ref.~\cite{Huang_etal20} on the model output to further boost accuracy. We implement C\&S  within the same framework as SAR  since C\&S involves iterative propagation of messages throughout the graph that is similar to a GNN layer. C\&S has no trainable parameters, though, and no backward pass.

We use the metis library~\cite{Karypis_etal97} to partition the graphs across the machines. Metis yields a roughly equal number of nodes in each partition which helps to balance the load and memory consumption across all machines. It also minimizes the number of edges that cross the partition boundaries which minimizes the volume of communication between the machines during training.

\begin{figure}[h]
  \centering
  \begin{subfigure}{0.35\textwidth}
    \includegraphics[width = \textwidth]{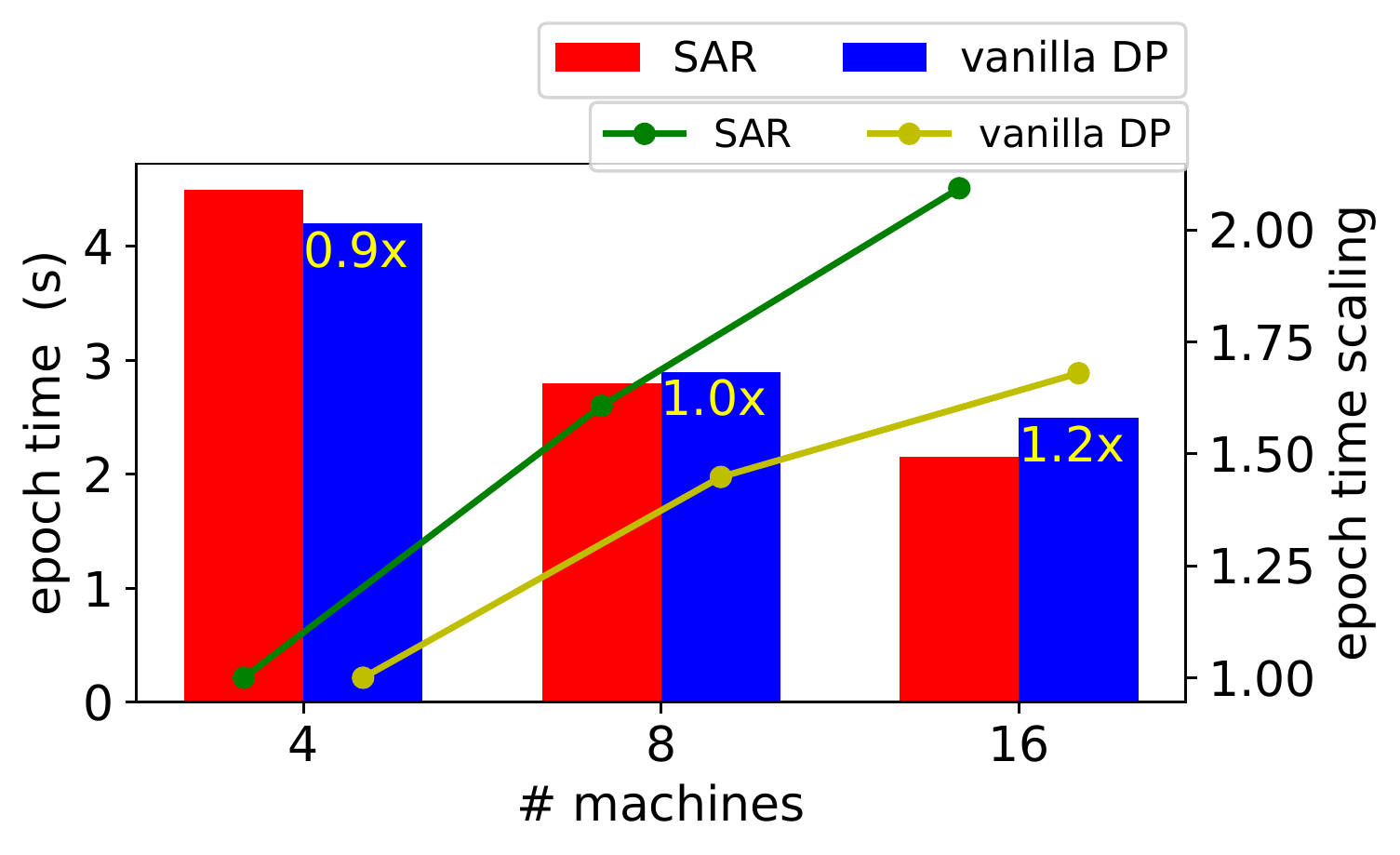} 
    \subcaption{}
    \label{fig:prods_sage_rt}
  \end{subfigure}
  \\
  \begin{subfigure}{0.35\textwidth}
    \includegraphics[width = \textwidth]{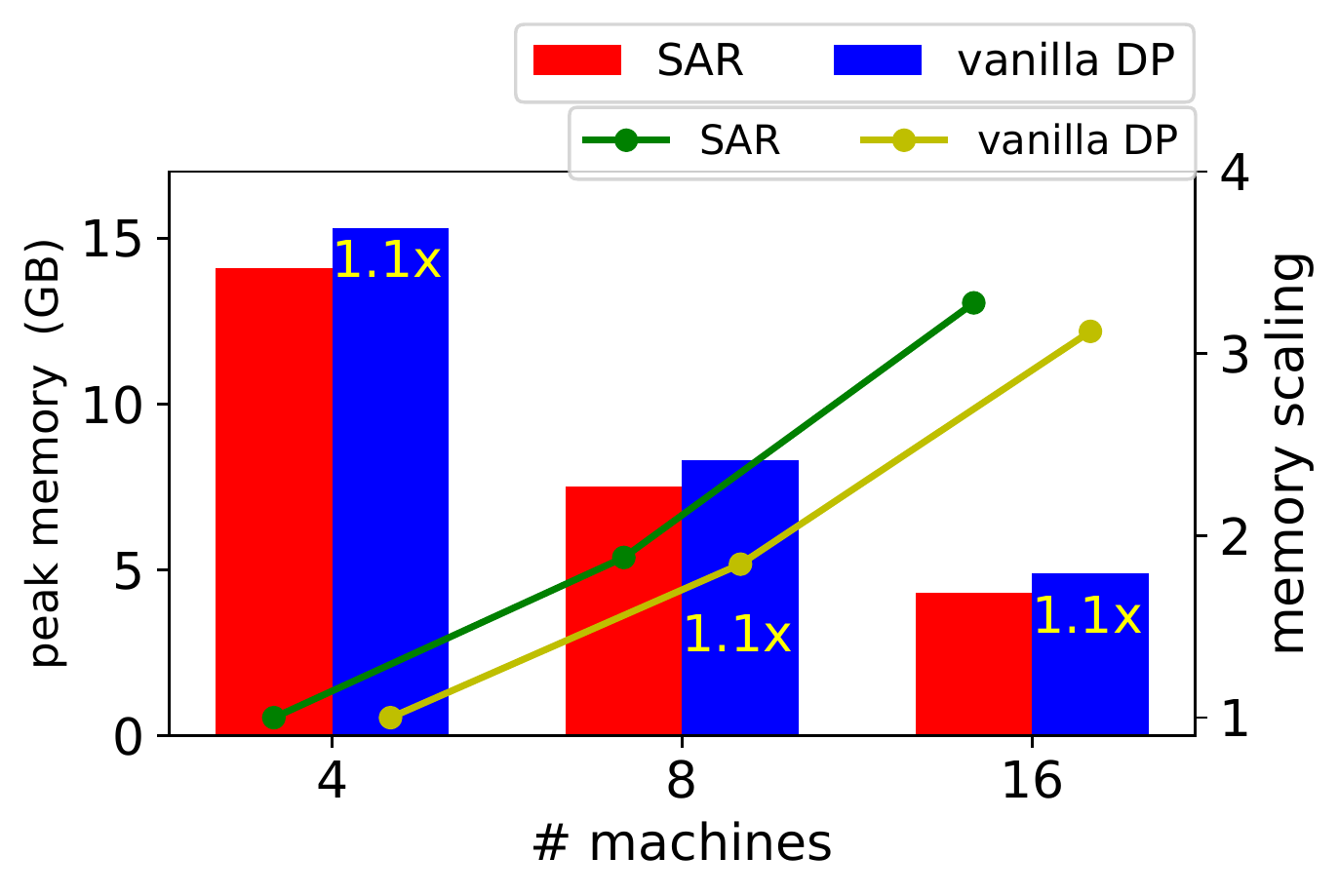} 
    \subcaption{}    
    \label{fig:prods_sage_memory}
  \end{subfigure}
  
  \caption{Epoch time and peak memory consumption per worker when training a 3-layer GraphSage network on ogbn-products.}
  \label{fig:prods_sage}
\end{figure}

\begin{figure}[h]
  \centering
  \begin{subfigure}{0.35\textwidth}
    \includegraphics[width = \textwidth]{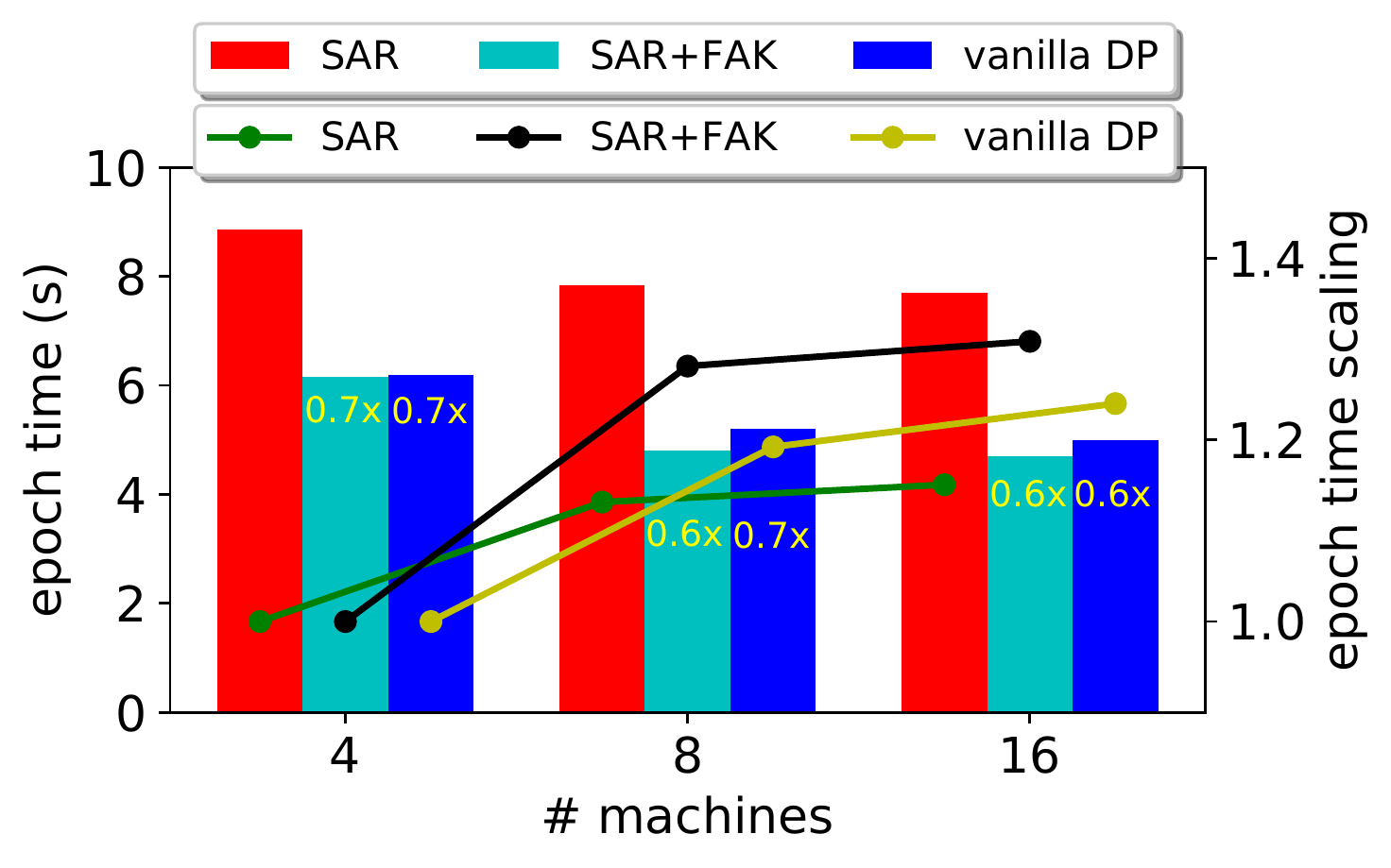} 
    \subcaption{}
    \label{fig:prods_gat_rt}
  \end{subfigure}
  \quad
  \begin{subfigure}{0.35\textwidth}
    \includegraphics[width = \textwidth]{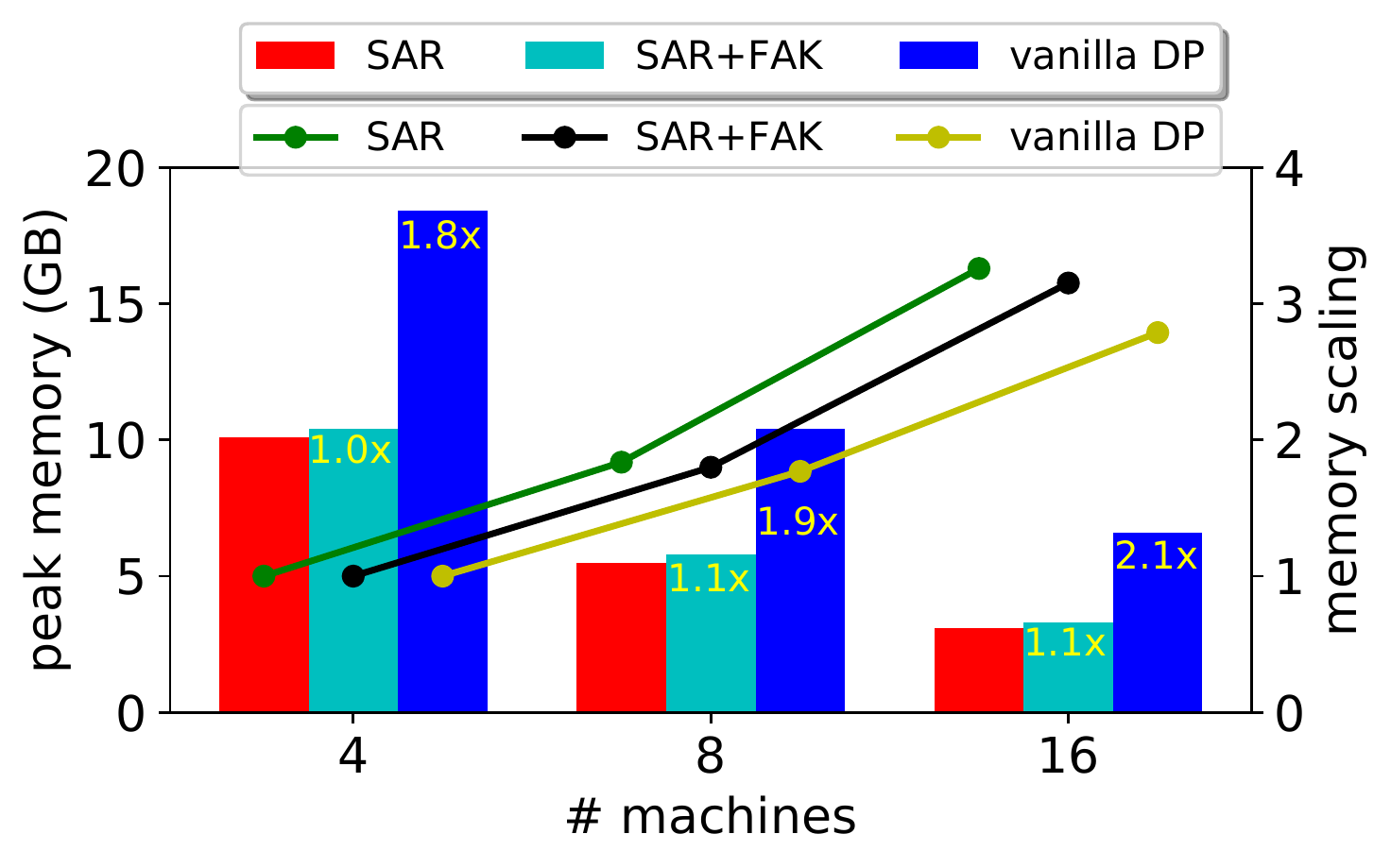} 
    \subcaption{}    
    \label{fig:prods_gat_memory}
  \end{subfigure}
  
  \caption{Epoch time and peak memory consumption per worker when training a 3-layer GAT network on ogbn-products. The scaling numbers in the bars are relative to SAR. }
  \label{fig:prods_gat}
\end{figure}

\paragraph{ogbn-products}: We partition the ogbn-products graph across 4, 8, and 16 machines. Figures~\ref{fig:prods_sage_rt} and~\ref{fig:prods_sage_memory} show the epoch time and peak memory consumption when using SAR and when using vanilla domain parallel training to train a GraphSage GNN. For GraphSage, SAR incurs no extra communication overhead. SAR has a slight run-time advantage as the number of partitions increases and is more memory-efficient.  Figures~\ref{fig:prods_gat_rt} and~\ref{fig:prods_gat_memory} show the GAT training results. Plain SAR is significantly slower than domain parallel training as SAR incurs a communication overhead compared to domain parallel training by re-communicating the nodes' activations during the backward pass. Augmenting SAR with our fused attention kernel (FAK), however, significantly accelerates training. Even though SAR+FAK still incurs a communication overhead compared to domain parallel training, it is just as fast due to the more efficient kernels used (see Fig.~\ref{fig:FAK}), and the fact that communication is not the main bottleneck. The memory consumption of SAR and SAR+FAK is significantly lower than domain parallel training, especially as we increase the number of workers. 

\begin{figure}[h]
  \centering
  \begin{subfigure}{0.35\textwidth}
    \includegraphics[width = \textwidth]{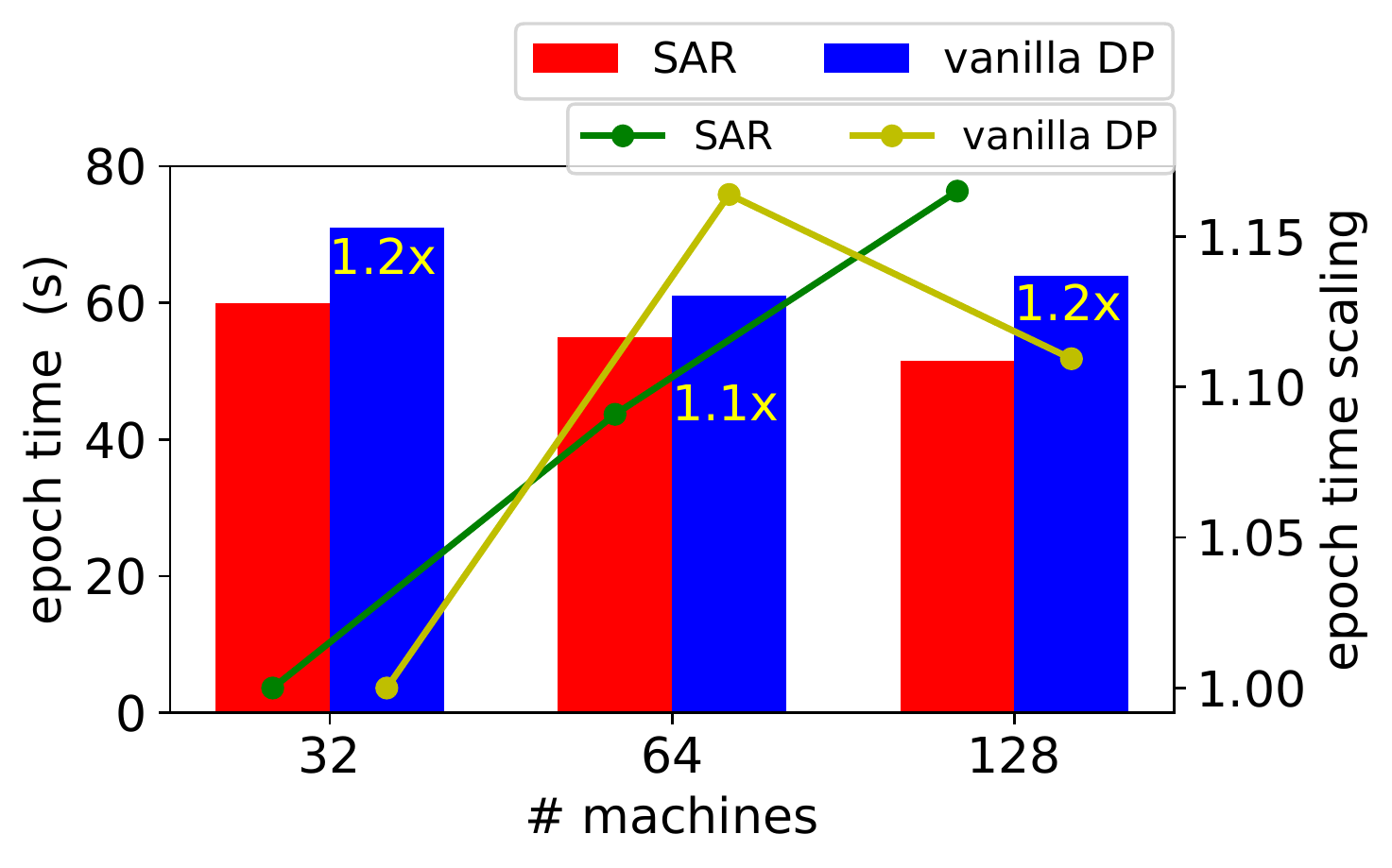} 
    \subcaption{}
    \label{fig:papers_sage_rt}
  \end{subfigure}
  \\
  \begin{subfigure}{0.35\textwidth}
    \includegraphics[width = \textwidth]{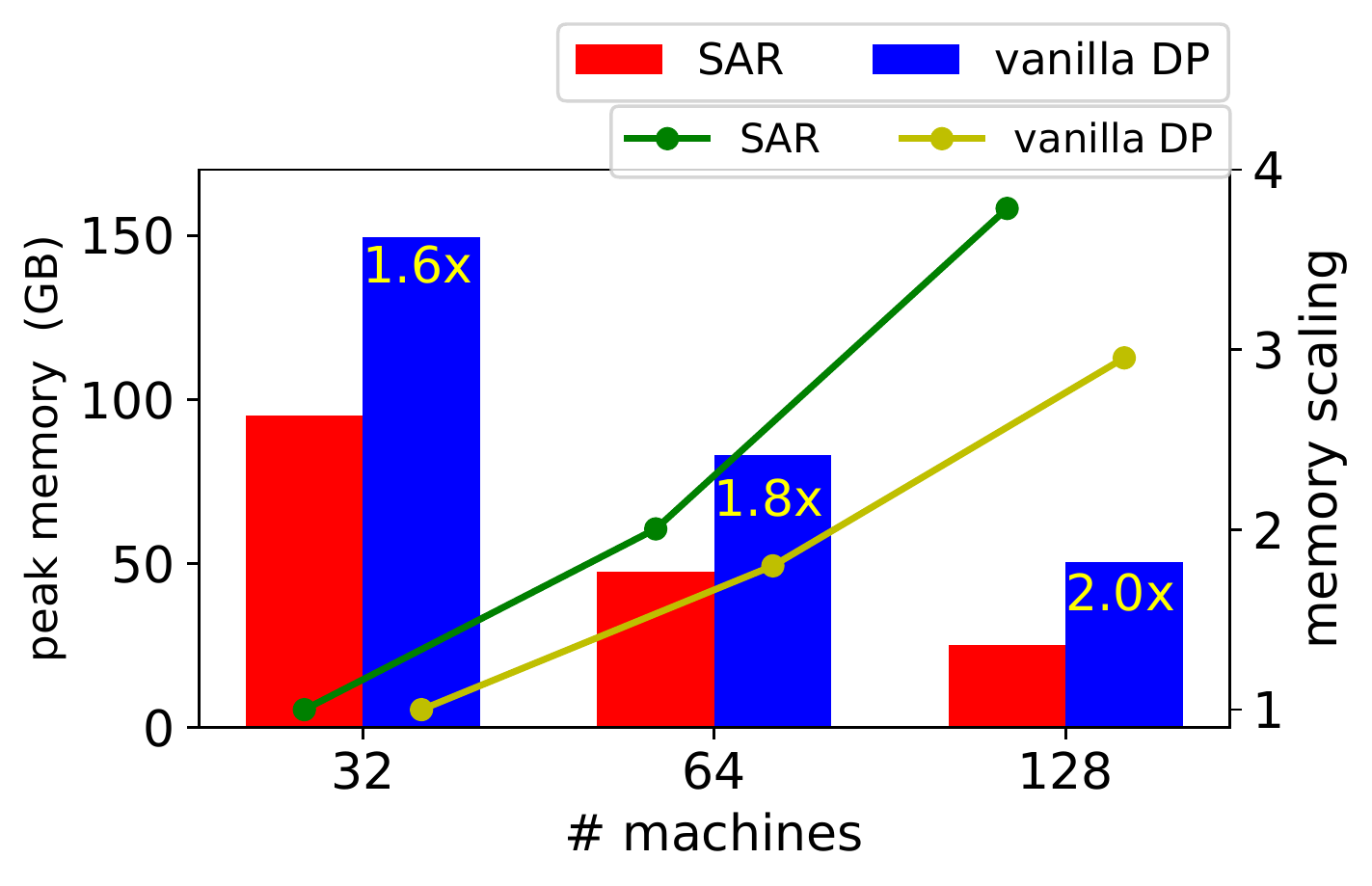} 
    \subcaption{}    
    \label{fig:papers_sage_memory}
  \end{subfigure}
  
  \caption{Epoch time and peak memory consumption per worker when training a 3-layer GraphSage network on ogbn-papers100M.}
  \label{fig:papers_sage}
\end{figure}

\begin{figure}[h]
  \centering
  \begin{subfigure}{0.35\textwidth}
    \includegraphics[width = \textwidth]{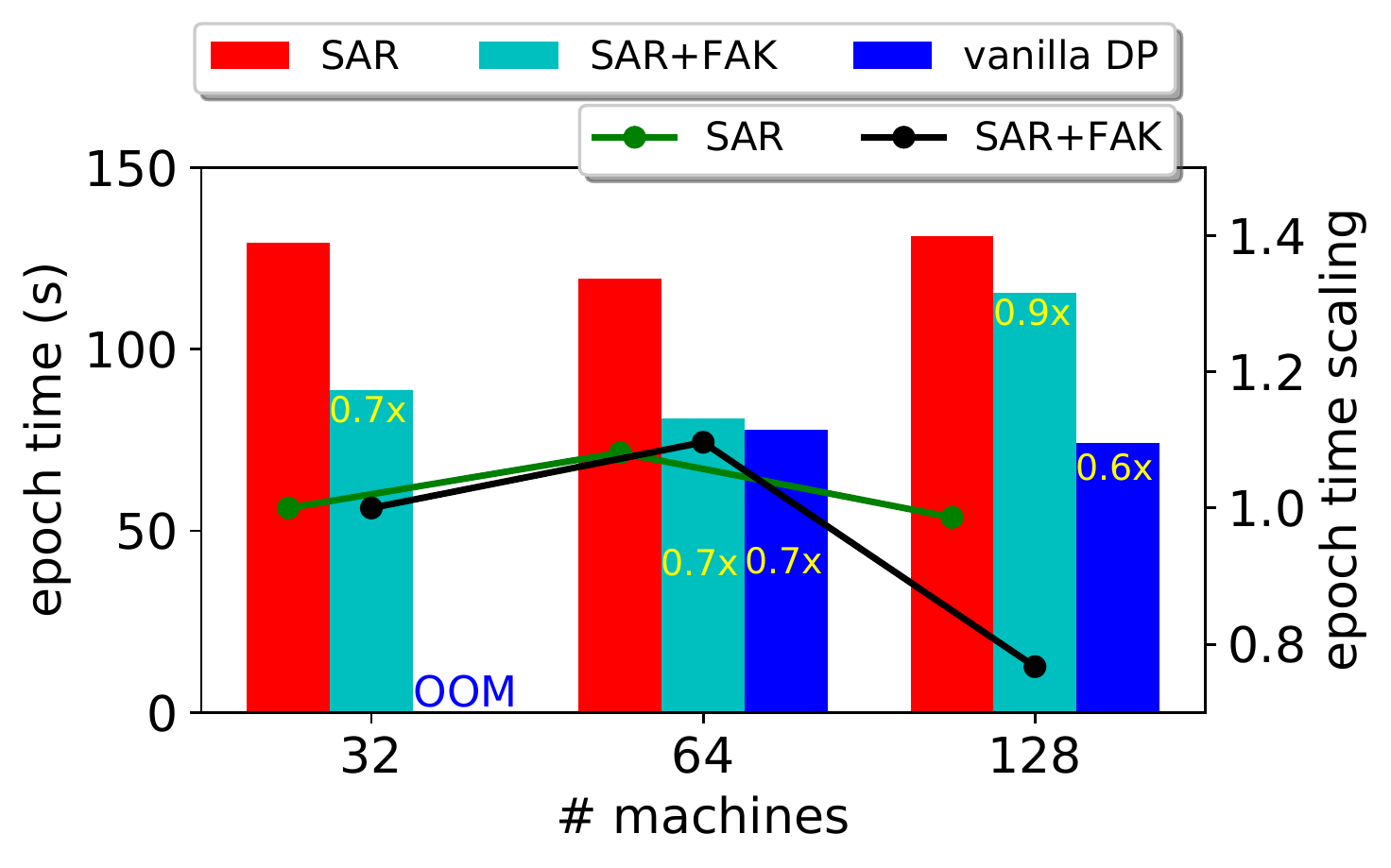} 
    \subcaption{}
    \label{fig:papers_gat_rt}
  \end{subfigure}
  \quad
  \begin{subfigure}{0.35\textwidth}
    \includegraphics[width = \textwidth]{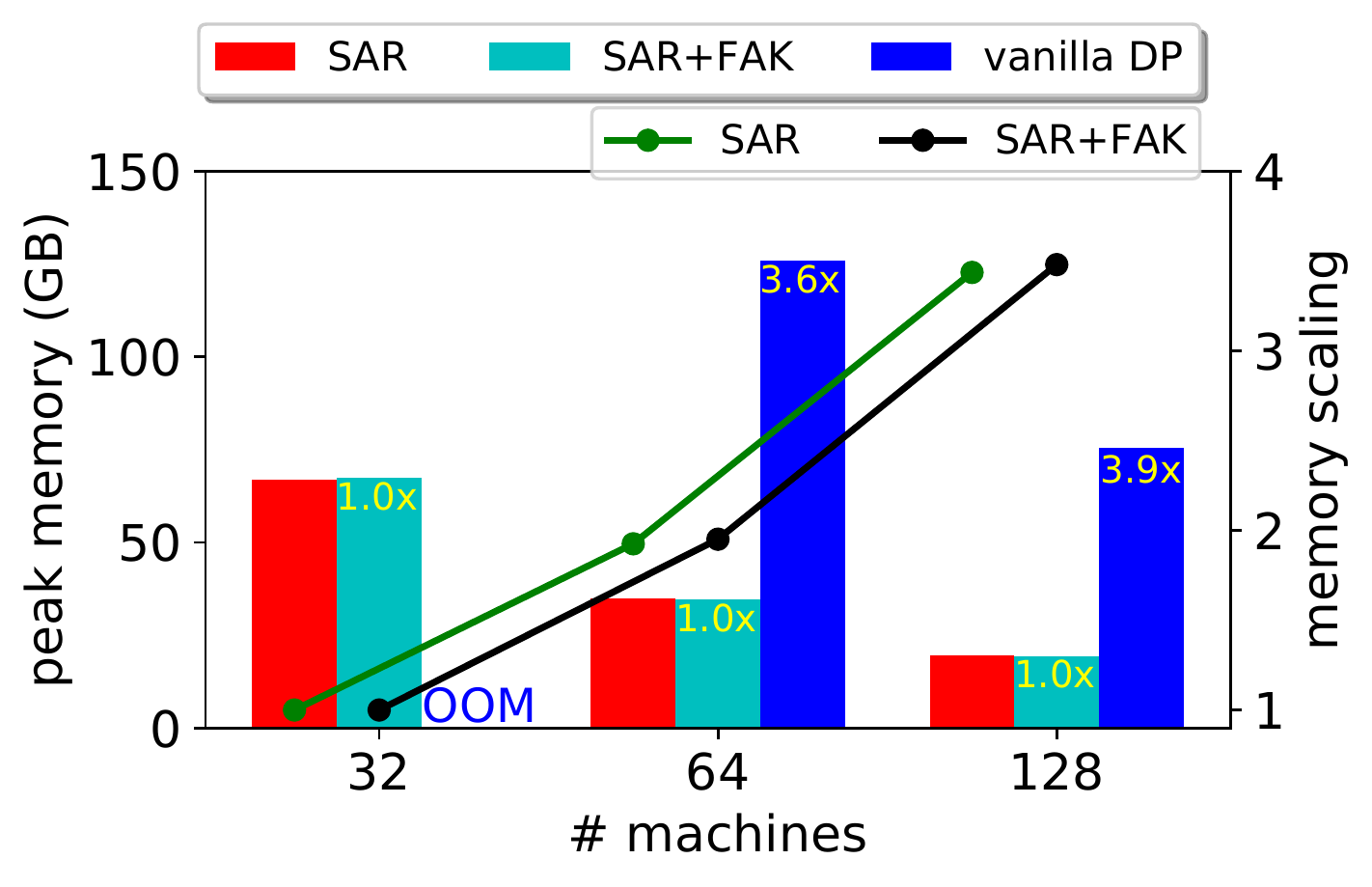} 
    \subcaption{}    
    \label{fig:papers_gat_memory}
  \end{subfigure}
  
  \caption{Epoch time and peak memory consumption per worker when training a 3-layer GAT network on ogbn-papers100M. The scaling numbers in the bars are relative to SAR.}
  \label{fig:papers_gat}
\end{figure}

\paragraph{ogbn-papers100M}: We partition the ogbn-papers100M graph across 32, 64, and 128 machines. Figures~\ref{fig:papers_sage} and~\ref{fig:papers_gat} show the epoch times and peak memory consumption for GraphSage and GAT, respectively. The memory advantages of SAR and SAR+FAK versus domain parallel training are more pronounced for ogbn-papers100M. SAR and SAR+FAK are almost 4x more memory efficient than domain parallel when training GAT on 128 machines. Training GAT on 32 machines is not possible with domain parallel training due to out-of-memory (OOM) errors. However, the extra communication overhead of SAR and SAR+FAK on GAT adversely affects their run-time scaling behavior as we increase the number of machines. Going from 64 to 128 workers increases epoch run-time for SAR and SAR+FAK as training becomes communication-bound. Domain parallel training on GAT at 128 workers is thus significantly faster as it involves less communication. For GraphSage where the communication volume is the same for all methods, SAR has a slight run-time advantage. SAR can cut memory consumption by half when training the GraphSage network on 128 machines. Appendix~\ref{sec:mfgs} contains additional results that quantify the convergence speed of training. Appendix~\ref{sec:mfgs} also  describes some key optimizations that can greatly improve the run-time on ogbn-papers100M.

\section{Conclusions}
We presented SAR, a distributed memory-efficient scheme for training large-scale GNNs. SAR ensures the large GNN computational graph is never fully materialized at once at any machine. Instead SAR materializes the computational graph piece by piece to enable incremental backpropagation. SAR drastically reduces memory consumption for GNN models such as GAT whose computational graph includes many large intermediate variables. Since SAR re-calculates the intermediate variables (such as the attention coefficients) during the backward pass, we developed custom GAT kernels that do not save these coefficients to memory (since this memory will be cleared by SAR anyway), but instead calculate and use the attention coefficients on the fly. The reduced memory pressure improved the speed of the forward pass, while keeping the backward pass comparable to that of DGL. Our custom kernels significantly improve SAR's performance when training GAT GNNs.

With the proliferation of approximate large-scale GNN training methods, exact (full-batch) training is a natural but hard to obtain, and rarely provided, baseline. SAR makes it easier to obtain this exact GNN training baseline in order to assess the effect of various training approximations on GNN performance. The question of which of the three is better: full-batch training, sampling-based training~\cite{Hamilton_etal17,Zeng_etal19,Chiang_etal19}, or training using non-learnable message propagation~\cite{Rossi_etal20,Sun_Wu21,Klicpera_etal18,Wu_etal19} is beyond the scope of this paper. The answer depends on the task at hand, the available computational resources, and the desired accuracies. For example, epoch time in sampling-based methods will generally be faster than full-batch training as the communication volume in distributed sampling-based training is significantly less. However, our accuracy figures for ogbn-products for example are significantly better than those reported for distributed sampling-based training~\cite{Zheng_etal20}. Moreover, it is unclear how the input labels augmentation scheme we used~\cite{Shi_etal20} would work in a sampling-based method where labels cannot propagate throughout the whole graph. Similarly, networks like GAT depend on having the full neighborhood for each node in order to correctly normalize the attention coefficients, which would not be possible in a sampling-based approach. Similarly, GNN training methods based on non-learnable message propagation achieve state of the art results on several node classification tasks~\cite{Sun_Wu21}, and are significantly faster than full-batch or sampling-based training methods that use learnable messages. Yet, they are less general than traditional GNNs, and have not yet been successfully applied to link prediction or graph property prediction tasks, tasks at which traditional GNNs perform extremely well~\cite{Zhang_etal20,Li_etal20}.

While we described SAR primarily in the context of full-batch GNN training, the underlying idea is generally applicable to any domain-parallel training situation where the input is partitioned across multiple workers, and each worker's output depends on parts of the inputs to other workers. One example is spatially-parallel convolutional neural networks~\cite{Dryden_etal19,Jin_etal18}. We plan on open-sourcing our general implementation of SAR to enable memory-efficient distributed full-batch training of GNNs, and to provide the tools needed to apply SAR to other distributed training problems.

\appendix
\section{SAR and Relational GCNs}
\label{sec:rgcn}
SAR is a general distributed GNN training method that is also applicable to heterogeneous graphs. We use SAR to train a Relational-GCN (R-GCN)~\cite{Schlichtkrull_etal18} on ogbn-mag, which is a heterogeneous graph dataset with 1.9 million nodes, 21 million edges, and 4 edge types. A R-GCN layer is described by:
\begin{equation}
h_i^{l+1} = \sigma(\sum\limits_{r \in {\mathcal R}} \sum\limits_{j \in {\mathcal N}^r_i}\frac{1}{|{\mathcal N}^r_i|} {\bf W}_r^{l} h_j^{l}),
\end{equation}
where $h_i^l$ is the feature vector of node $i$ at layer $l$, ${\mathcal R}$ is the set of all relations, ${\mathcal N}_i^r$ is the set of neighbors of node $i$ which connect to node $i$ using the relation $r$. ${\bf W}_r^{l}$ are relation-specific learnable parameters in layer $l$. 

We can use several methods to reduce the number of learnable parameters (the ${\bf W}_r^{l}$s). One common method is to represent these parameters as linear combinations of few shared basis tensors:
\begin{equation}
{\bf W}_r^l = \sum\limits_{b = 1}^B a_{rb}^l{\bf V}_b^l,
\end{equation}
where $B$ is the number of basis tensors, ${\bf V}_b^l$ are the basis tensors, and $a_{rb}$ are relation-specific coefficients.

Note that the aggregation operator of R-GCN has learnable parameters, and backpropagating gradients to these learnable parameters (the ${\bf W}_r^{l}$s) requires the values of the layer's input features (${\bf h}^{l}$). As described in Section~\ref{sec:sar}, SAR would thus need to re-fetch remotely stored node features during the backward pass to correctly calculate the gradients (case 2 in Section~\ref{sec:sar}). 

\begin{figure}[h]
  \centering
  \begin{subfigure}{0.35\textwidth}
    \includegraphics[width = \textwidth]{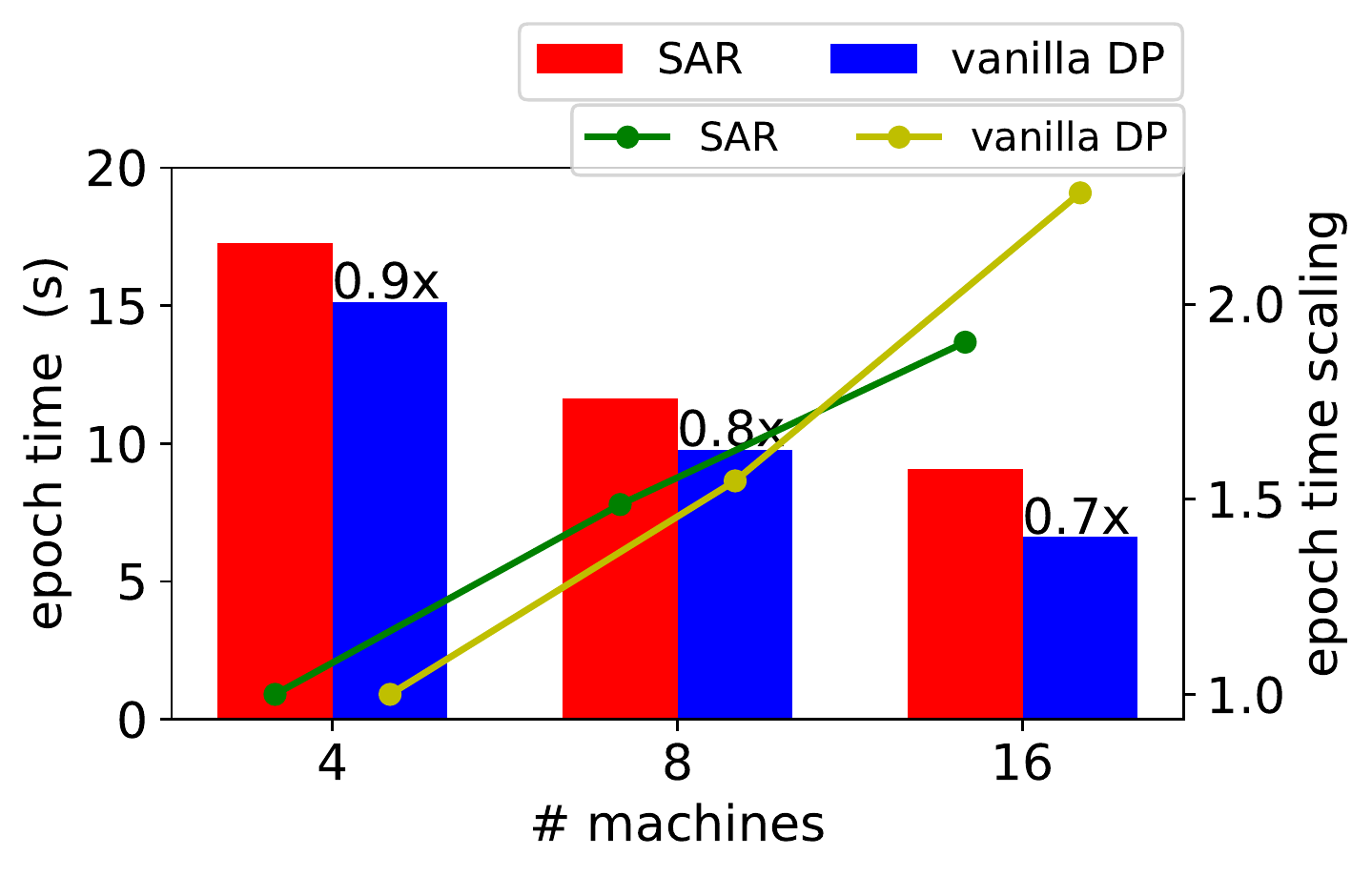} 
    \subcaption{}\label{fig:mag_rt}
  \end{subfigure}
  \\
  \begin{subfigure}{0.35\textwidth}
    \includegraphics[width = \textwidth]{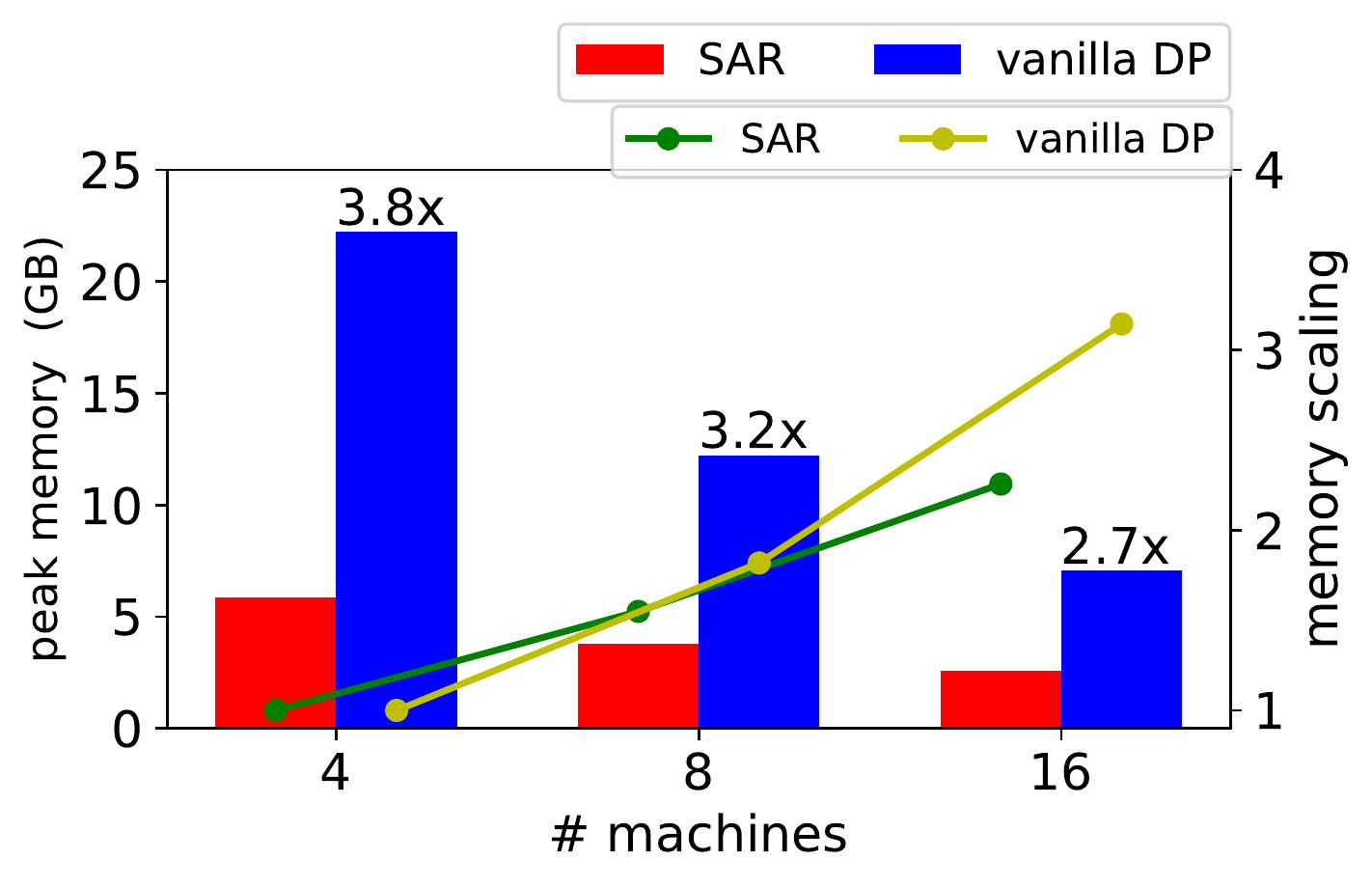} 
    \subcaption{}\label{fig:mag_memory}
  \end{subfigure}
  
  \caption{Epoch time and peak memory consumption per worker when training a 3-layer R-GCN network on ogbn-mag.}\label{fig:mag_sage}
\end{figure}

We directly use DGL's RelGraphConv layer to implement a 3-layer R-GCN network with hidden layer sizes of 256. Figure~\ref{fig:mag_rt} shows the run-time scaling behavior. Since SAR incurs extra communication and computation overhead to re-fetch remote features and reconstruct the computational graph during the backward pass, its epoch run-time lags behind vanilla Domain Parallel (DP) training. However, as shown in Fig.\ref{fig:mag_memory}, SAR is significantly more memory efficient, requiring only $~26\%$ to $~37\%$ of the memory needed by vanilla DP training.

\section{Convergence Speed of Full-batch Training}
\label{sec:mfgs}
A full-batch GNN training method only updates the GNN parameters at the end of each epoch/iteration. This is in contrast to sampling-based methods which may update the parameters after every mini-batch of sampled graphs. More frequent parameter updates is one potential advantage of sampling-based GNN training methods, even though these updates might be biased due to the incomplete neighborhoods in sampled graphs.

\begin{figure}[h]
  \centering
  \includegraphics[width = 0.4\textwidth]{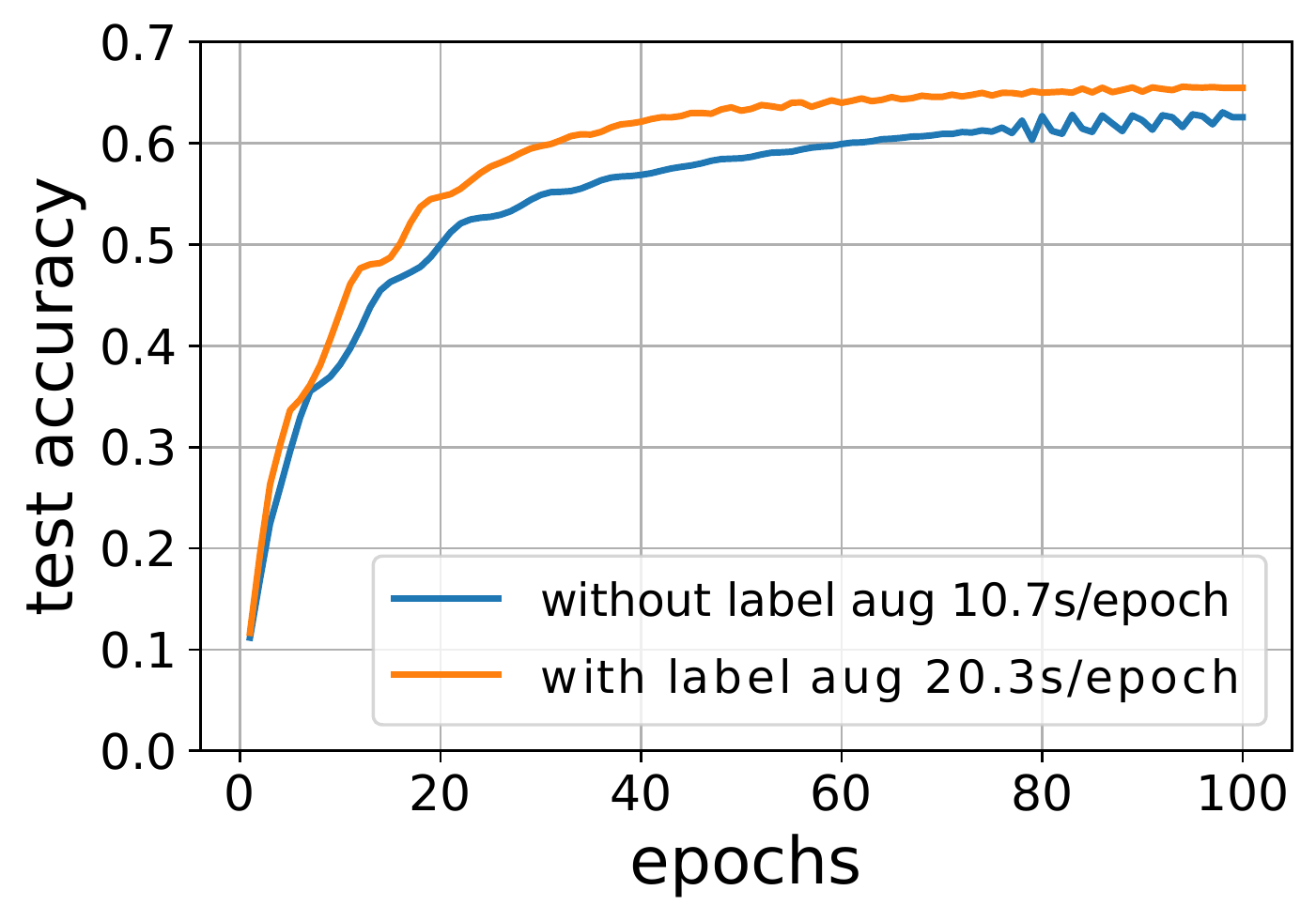} 
  \caption{Training curve when using SAR to train a 3-layer GraphSage network on ogbn-papers100M. We trained on 32 2-socket machines equipped with Intel's 36-core IceLake processors and connected using Infiniband HDR 200Gb/s interconnects.}\label{fig:papers_mfg}
\end{figure}

\begin{figure}[h]
  \centering
  \includegraphics[width = 0.4\textwidth]{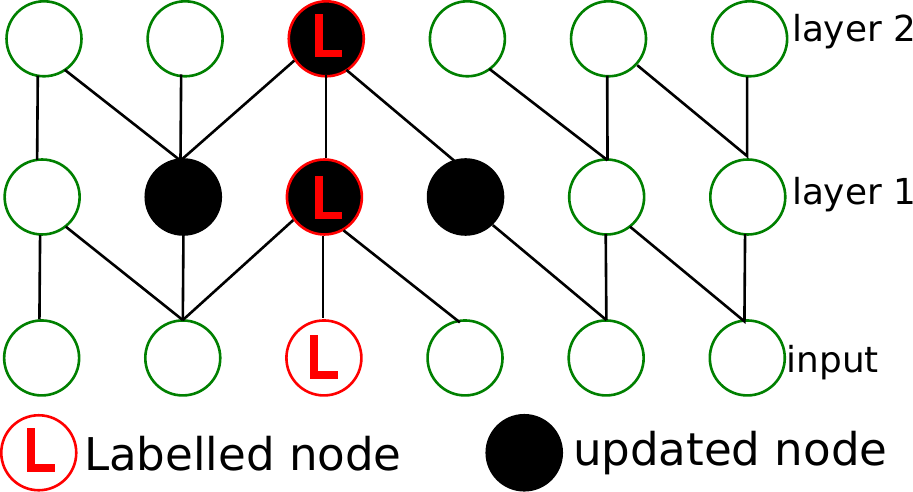} 
  \caption{Example graph with one labeled node showing the nodes that need to be updated at each layer of a 2-layer GNN.}\label{fig:mfg}
\end{figure}

We show that full-batch training  converges after a reasonable number of epochs/weight updates. Figure~\ref{fig:papers_mfg} shows the training curve for full-batch training on ogbn-papers100M. We used a 3-layer GraphSage network with hidden layer size of 256 and batch normalization between all layers. We ran experiments with and without label augmentation~\cite{Shi_etal20}. As shown in Fig.~\ref{fig:papers_mfg}, training practically converges after 100 epochs, 

Full-batch GNN training methods typically compute the feature vectors of all nodes in the graph at each layer. Current sampling-based methods, however, usually construct what is known as Message Flow Graphs (MFGs). In node classification tasks, MFGs allow sampling-based method to only compute the node features at each layer which will affect the classification output at the labeled nodes. This is illustrated in Fig.~\ref{fig:mfg} which shows an example graph with 6 nodes, 10 edges, and only one labeled node. The figure shows the message passing edges in a 2-layer GNN. Since the gradient only backpropagtes from the labeled node, we only need to update the black nodes at each layer, as these are the only nodes that affect the output.

We used DGL's MFGs to optimize SAR's computation, and avoid updating unnecessary nodes at each layer. As shown in Fig.~\ref{fig:papers_mfg}, this pushes the epoch run-time down to 10.7 seconds for a vanilla GraphSage network and 20.3 seconds for a GraphSage network that uses label augmentation.

\end{document}